\documentclass[conference,compsoc]{IEEEtran}
\ifCLASSOPTIONcompsoc
  \usepackage[nocompress]{cite}
\else
  \usepackage{cite}
\fi
\usepackage{amsmath}

\ifCLASSINFOpdf
\else
\fi
\usepackage{graphicx}
\usepackage{enumitem}
\newlist{tabitem}{itemize}{1}
\setlist[tabitem]{wide=0pt, nosep, leftmargin= * ,label=\textbullet,after=\vspace{-\baselineskip},before=\vspace{-0.6\baselineskip}}

\begin{document}
%

\title{Breast Cancer Segmentation using Attention-based Convolutional Network and Explainable AI}


%


\author{\IEEEauthorblockN{
\large Jai Vardhan \IEEEauthorrefmark{1}, Taraka Satya Krishna Teja Malisetti\IEEEauthorrefmark{2}\\
\IEEEauthorblockA{\IEEEauthorrefmark{1}\normalsize Department of Computer Science, IIIT Naya Raipur, India}
\IEEEauthorblockA{\IEEEauthorrefmark{2}\normalsize Master of Science in Business Analytics, University of Louisville, USA}
\normalsize {e-mail:  jai20100@iiitnr.edu.in, tarakteja.malisetti@gmail.com}
}}

\maketitle

\begin{abstract}
One of the most hazardous diseases for people is cancer, yet no long-term treatment is currently available. One of the most typical cancers is breast cancer. Early detection of breast cancer is the goal of screening mammography, as this allows for more effective therapy. High false positives and negative rates impact the interpretation of mammograms despite the existence of screening programmes worldwide. With an estimated 2.3 million new cases or 11.7\% of all cancer cases, BC is expected to overtake lung cancer as the world's most prominent cause of cancer incidence in 2020. Early detection of Breast Cancer(BC) is the goal of screening mammography, as this allows for more effective therapy. The interpretation of mammograms is impacted by high false positives and negative rates despite the existence of screening programmes worldwide. There have been advances in several technologies that have helped reduce the mortality rate from this illness. Still, early discovery is the most effective way to stop the progress of the disease, amputation of the breast, and death. Infrared cameras with outstanding resolution and sensitivity are used in thermography, a promising tool for early detection. It is predicted that using thermal imaging in conjunction with artificial intelligence (AI), would produce outstanding predictability levels for the early detection of breast cancer. The current work utlizes attention-based convolution neural network for segmentation. When compared to existing works the current system is more precise and faster in detecting and classifying the BC diseases. Furthermore, this framework comprises image enhancement, and cancer segmentation with explainable AI (XAI). A transformer-attention-based \cite{9042231} convolutional architecture (UNet) is proposed for fault identification. Moreover, to analyze the region of bias and weakness of the UNet architecture with IRT images, the Gradient-weighted Class Activation Mapping (Grad-CAM) is performed. The transcendence of the proposed architecture is verified in comparison with existing state-of-the-art deep learning frameworks for the BC dataset.
\end{abstract}
\begin{IEEEkeywords}
Breast Cancer Detection, Image classification, Vision
Transformers, Explainable AI, Convolutional Neural Networks
\end{IEEEkeywords}

\section{Introduction}
\subsection{History of Breast Cancer(BC) Detection}
Breast cancer, a disease that has plagued humanity for centuries, has witnessed significant advancements in understanding, diagnosis, and treatment. Our knowledge about breast cancer can be traced back to ancient times, specifically to the Edwin Smith Surgical Papyrus, which dates back to 3,000-2,500 B.C. This historical document, attributed to the esteemed Egyptian physician-architect Imhotep, contains detailed accounts of breast cancer cases. Interestingly, it describes specific characteristics believed to render the disease incurable, such as a breast that felt cold to the touch, appeared bulging, and exhibited widespread growth. The concept of hormonal involvement in the disease emerged in the quest to comprehend the underlying factors contributing to breast cancer. Observations revealed that breast cancer tends to be more severe in younger women, leading to the hypothesis that hormones might play a significant role. Notably, these ideas predated the discovery of estrogen receptors by Jensen in 1967. In 1906, a pioneering Scottish surgeon named Beatson \cite{singh2022video} introduced the concept of endocrine surgery by performing oophorectomy and adrenalectomy, essentially achieving castration, as a treatment for breast cancer. However, as medical understanding progressed, more refined approaches emerged. Extreme measures gave way to using estrogen receptor modulators, luteinizing hormone-releasing agonists, and aromatase inhibitors, which proved more effective and less invasive. The development of diagnostic techniques has been crucial in the fight against breast cancer. In 1913, radiographs, or X-rays, were first utilized to examine breast cancer patients, offering valuable insights into the disease. Building upon this progress, a German surgeon named Salmon conducted a study involving 3,000 patients, contributing to the growing knowledge base on breast cancer. Another significant milestone came in 1951 when ultrasound research in breast cancer detection commenced. The objective was to develop a technique to identify breast tumours and determine their malignancy. In 1952, with the sponsorship of 21 cases from other research, successful detection of breast cancer using ultrasound was achieved, leading to further evaluation of hospital ultrasonography \cite{article3} equipment. By 1954, ultrasound had firmly established itself as a promising method for identifying breast cancer. Advancements in ultrasound technology continued throughout the 1960s. The internal architecture of ultrasound systems was improved, and detection techniques were refined, enhancing the accuracy and reliability of breast cancer diagnosis. Innovations such as immersing breasts in controlled-temperature water during early pregnancy to aid tumour identification were explored.
In 1982, thermal imaging, utilizing infrared technology, received approval from the U.S. Food and Drug Administration (FDA) as a diagnostic tool for breast cancer. This non-invasive technique involved capturing images that depict variations in temperature within the breast, potentially highlighting areas of concern. A landmark study conducted in 1996 compared thermal imaging with traditional X-rays for breast cancer diagnosis. The study successfully demonstrated that thermal imaging could effectively identify breast cancer in a patient, even when X-ray methods failed to do so. Recently authors in \cite{vardhan2023detection} used deep learning techniques to detect diseased leaves from drone-captured images that could be equipped in our case. These advancements in the understanding and diagnosis of breast cancer have significantly contributed to improving patient outcomes and shaping the field of oncology. From ancient medical texts to modern technological innovations, the ongoing pursuit of knowledge and innovation continues to propel the fight against breast cancer, offering hope for early detection, personalized treatment, and improved survival rates.

\begin{table*}[t]
\begin{center}

\begin{tabular}{ |p{4cm}|p{6cm}|p{6cm}|  }
\hline
\textbf{Method} & \textbf{Advantages} & \textbf{Disadvantages} \\
\hline
Edge-Based & 
\begin{tabitem}
    \item Complex in the existence of extensive or poorly determined edges
\end{tabitem}
& 
\begin{tabitem}
    \item Quick
    \item Performs skillfully with high-contrast photos. 
\end{tabitem}
\tabularnewline
\hline
Region Based
& 
\begin{tabitem}
    \item Laborious
    \item Excessive segmentation
    \item Demanding the choice of a seed point
\end{tabitem}
& 
\begin{tabitem}
    \item For Uniformity, Region Growing is preferred.
    \item Split and Merge segmentation
    \item Watershed- delivers closed boundaries.
\end{tabitem}
\tabularnewline
\hline
Clustering
& 
\begin{tabitem}
    \item Costly
    \item Prone to outliers and initial groupings
\end{tabitem}
& 
\begin{tabitem}
    \item Performs satisfactorily for overlapped image data.
\end{tabitem}
\tabularnewline
\hline
Thresholding 
& 
\begin{tabitem}
    \item Do not operate satisfactorily with noise and low contrast images
    \item Improper results in larger segmentations
\end{tabitem}
& 
\begin{tabitem}
    \item Uncomplicated
    \item Rapid computation
\end{tabitem}
\tabularnewline
\hline
Energy function Based \cite{Said2018}
& 
\begin{tabitem}
    \item Susceptible to snake initialization
\end{tabitem}
& 
\begin{tabitem}
    \item Adjustable
    \item Need minor computation
\end{tabitem}
\tabularnewline
\hline
\end{tabular}

\end{center}
\caption{Comprehensive Review of BC Segmentation Techniques}
\label{table:1}
\end{table*}

\subsection{ Motivation for automated Diagnostic systems}
The computer vision and machine learning based mechanisms are highly used in classification and detection tasks in several industries including security \cite{singh2022video}, robotics \cite{krishna2022epersist}, Internet of Things \cite{reddy2022dscout}. In medical domain, computer vision is used for disease identification \cite{krishna2023lesionaid} and patient aiding \cite{nemani2022deep}.  Automated diagnostics improve evidence-based medicine, enhance treatment quality, encourage wellness, enable early illness identification, and lower total health care costs. Automation and technological advancements have improved tests' usability and accuracy, resulting in reports that are more accurate and timely. The requirement for breast cancer detection by automated diagnostic methods \cite{8930504} increased because so many people make mistakes while evaluating and identifying breast cancer.

\subsection{Research Challenges}
\subsubsection{Dataset}  
Breast Cancer Dataset images are challenging to analyze due to low image quality, noise, and varying viewpoints due to the handheld nature of the sensor. 
    \begin{itemize}
        \item Single Private Dataset
        \item Less number of Images
        \item Existence of Class Imbalance
    \end{itemize}

\subsubsection{Extraction of Region Of Interest} 
We could not employ any object detection algorithms in order to extract the necessary features. So, segmentation based methodologies are preferred.

\section{Related Works}
Segmentation is an essential image processing procedure, but it can be challenging. Segmentation separates a picture into numerous differentiable regions with different properties. Indeed, it is performed to remove a target from the background. The following techniques are the most standard and familiar image segmentation procedures: thresholding, clustering, region-based segmentation, energy function-based technique, and edge-based segmentation are mentioned with their advantages and disadvantages in Table \ref{table:1}. 

\subsection{Edge-based Segmentation}
Generally, the edge-based technique establishes the borders between areas to segregate them by employing the discontinuity in degree values of pixel between them. These methodologies are rapid in real-time application implementations. It detects edge features via color variation which can be perceived as a  black background with white lines. The Log, Roberts, Sobel, Prewitt, and Zero-cross filters can be applied to images for the same task. Nevertheless, the output quality is often inadequate for high-level computation.

\subsection{Region-based Segmentation}
The intensity levels of nearby pixels influence region-based segmentation. The colour, grey level, shape and texture of the visual dictated the homogeneity of the segmentation zones. To do so, the picture is broken into smaller sub-regions. The machine was needed to grasp and differentiate between desired and undesirable regions by utilising the information from pixel intensity, specified areas, and density. The median range of intensity levels between 1 (white) and 0 (black) can be used to separate a picture from its surrounding environment \cite{article} and fragment it. Density Slice represents the intensity distribution for the Region of Interest (ROI). After a picture is segmented, every pixel is reallocated based on whether the picture's intensity count is 1 or 0.

\subsection{Thresholding Methods}
Thresholding is the most basic picture segmentation approach. Thresholds are classified into two types: Bi-level thresholding partitions photos into only two categories, while Multilevel thresholding (MT) splits pictures into more than two categories.  Metaheuristic-dependent procedures such as ABC, BFO, and PSO, on the other hand, have indeed been utilized to estimate optimum MT values.
\subsection{Energy function based Segmentation}
The general Snakes procedure was presented by Kass et al. [12]; in
this procedure, a turn becomes beneath some energy till it arrests at
the border. The turn drives to undervalue the energy. This
procedure utilises a framework in which regional minima of energy
function has a collection of solutions. By counting functional energy periods to the depreciation, the user can drive the model out of regional minima towards the preferred solution. The consequence is an operational prototype that descends\cite{HASAN2021} into the preferred explanation when positioned around it. Kass's snake pinnacle is dynamic, always underestimates its exuberance procedure, and displays an engaged demeanour. This method is recognised under the class of parametric Dynamic silhouette models. Active contour energy-based methods can be
classified as parametric and geometric. The enthusiastic silhouette procedure uses parameterised turns to illustrate the shapes.
\section{Methodology}
Two subsections are included in this section. The pre-processing section provides different processes that are applied to the picture data, such as frame conversion and image scaling. The U-Net architecture is used to do further processing on the pre-processed pictures. On certain difficult benchmark datasets that are frequently used for breast cancer segmentation and classification, the study produced considerable results.

\begin{figure}[ht]
    \centering
    \includegraphics[width=\linewidth]{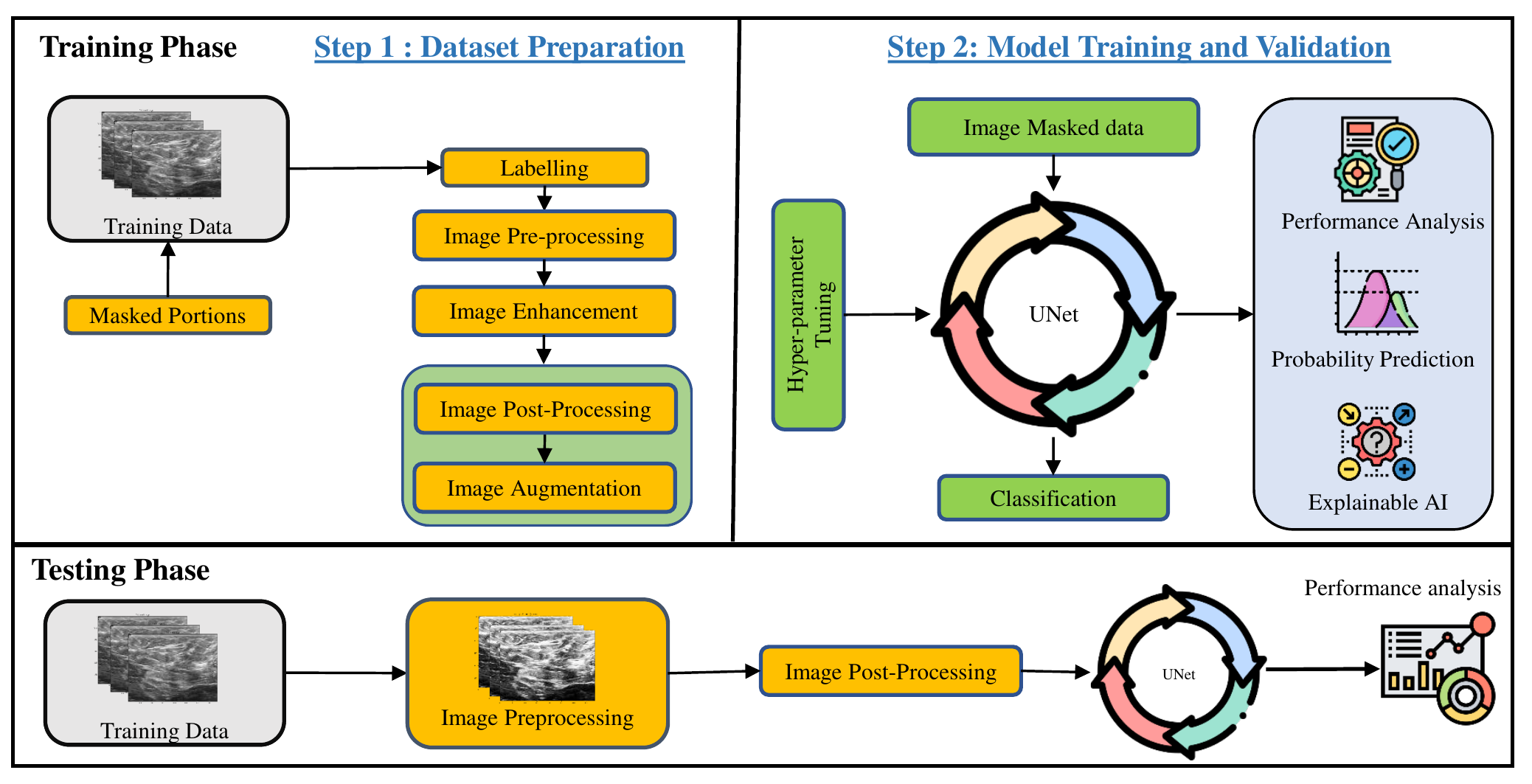}
    \caption{Conceptual Diagram of proposed architecture}
    \label{fig:architecture}
\end{figure}

\subsection{Dataset}
The data examines ultrasound scans used to diagnose breast cancer. Pictures from the Breast Ultrasound Dataset are divided into normal, benign, and malignant images. Breast ultrasound pictures can yield excellent results for categorising, detecting, and segmenting breast cancer when used in conjunction with machine learning. Women between the ages of 25 and 75 are represented by breast ultrasound pictures in the baseline data. These numbers were gathered in 2018. 600 female patients make up the total number of patients. With an average image size of 500*500 pixels, the collection comprises of 780 images. PNG format is used for the pictures. The original photos are displayed alongside the real-world photographs. Three classes—normal, benign, and malignant—are established for the photos.
\begin{figure}[ht]
    \centering
    \centerline{\includegraphics[scale = 0.4]{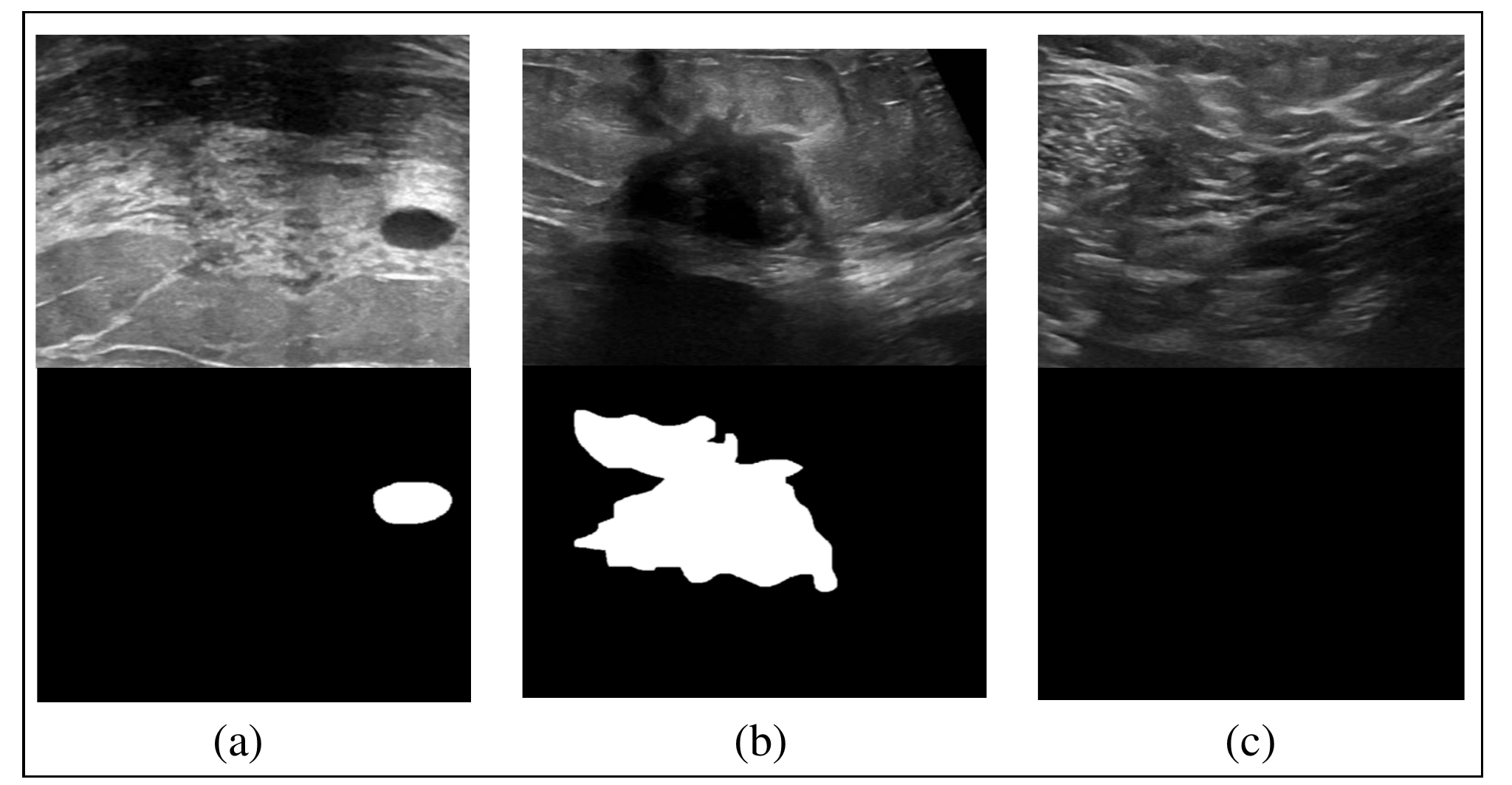}}
    \caption{Three classes of the dataset (a) Benign (b) Malignant (c) Normal }
    \label{fig:Aug
    }
\end{figure}

\subsection{Histogram Equalization: }
Industrial infrared images often contain industrial noise, such as dust, smoke and misty background objects. This industrial noise in an IR image is responsible for low performance in the predictive models. Apart from the noise, extracting the region of interest (ROI) is necessary to achieve better performances in predictive algorithms, reducing the redundant information and computational effort in pattern 

\begin{figure*}[ht]
    \centering
    \centerline{\includegraphics[width=\linewidth]{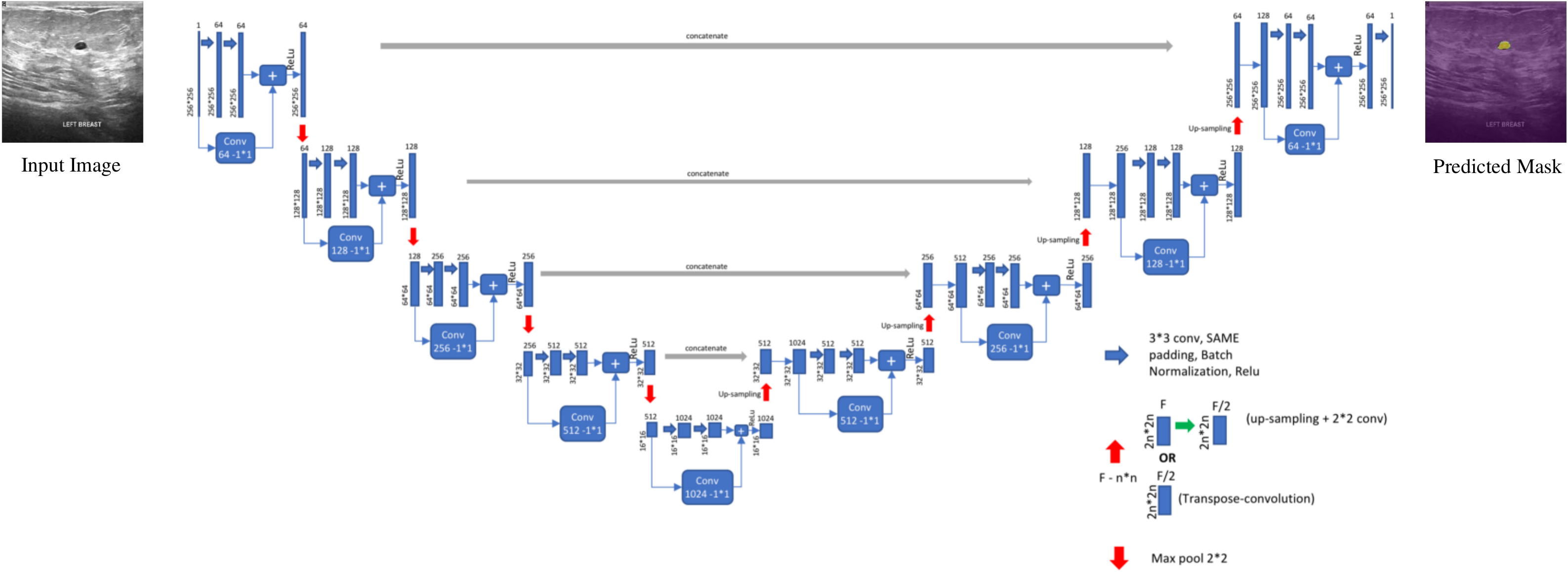}}
    \caption{U-Net Architecture Diagram}
    \label{fig:Arch}
\end{figure*}

recognition. Most existing architectures do not consider these essential factors; instead, they fit the unprocessed images into the predictive framework. Considering these factors, the proposed methodology enhances the image's quality and extracts the ROI by Contrast Limited Adaptive Histogram Equalization (CLAHE) \cite{9669234}. Unlike the existing architectures, this method reduces computational efforts and improves the overall performance of predictive models. CLAHE is an adaptation of adaptive histogram equalisation (AHE), which solves the problem of contrast over-amplification. Instead of processing complete image, CLAHE works with discrete sections of it called tiles. To eliminate the arbitrary borders, adjacent tiles are blended using bilinear interpolation. In contrast to other colour spaces, the acquired values of the quantitative metric characteristics, Entropy and RMS Contrast value, were significantly high. To highlight the region of interest and eliminate unnecessary background objects, CLAHE works efficiently with less computational time.

\subsection{Image Segmentation}
The process of grouping together portions of an image that belong to the same object class is known as image segmentation. This method is additionally known as "pixel-level classification." To put it another way, it entails dividing up images (or video frames) into a number of segments or objects. Categorizing pixels with semantic labels (semantic segmentation) or splitting specific objects are two ways to define image segmentation (instance segmentation). Semantic segmentation \cite{s21144833} labels every image pixel at the pixel level using a set of object categories. A single label is predicted for the whole picture or frame in image classification, which is typically a more challenging task. By locating and identifying all the items of interest in a picture, instance segmentation broadens the scope of semantic segmentation. A new category of image segmentation models with notable performance improvements has been introduced by deep learning models. Deep learning-based image segmentation models frequently attain the best accuracy rates on well-known benchmarks, leading to a paradigm shift in the industry. Image segmentation effectively separates the homogenous areas at a pixel-accurate scale, enhancing medical images like CT scans and MRI. Medical image processing has adapted a huge variety of segmentation techniques to identify the pixels of human organs.

\subsubsection{UNet for Image Segmentation}

We are able to conduct challenging tasks on computer vision datasets with high-quality results because to deep learning architectures like U-Net and CANet. Although the topic of computer vision is vast and has a wide range of interesting applications and challenges to solve, our focus in the upcoming articles will be on two architectures, U-Net and CANet, that are intended to address the problem of picture segmentation.
There are two paths in it: an expansive path and a contracting path. The contracting route follows to a convolutional network's standard architecture. Although U-Net represents a big breakthrough in deep learning, it is as important to comprehend the earlier approaches that were used to handle challenges of this nature. The sliding window technique, which easily won the EM segmentation competition at ISBI in 2012, was one of the key instances that came to an end.
It can observed why the design in the image is likely referred to as U-Net architecture by taking a quick glance at it. The following term is derived from the shape of the so-formed architecture, which is in the shape of a "U." We can tell that the network produced is a fully convolutional network just by looking at the structure and the many components used in the building of this architecture. They did not employ any additional layers, such as dense, flat, or layers of a similar nature. The graphic depiction demonstrates a route that initially contracts before expanding.

\begin{figure*}
    \centering
    \centerline{\includegraphics[width=\linewidth]{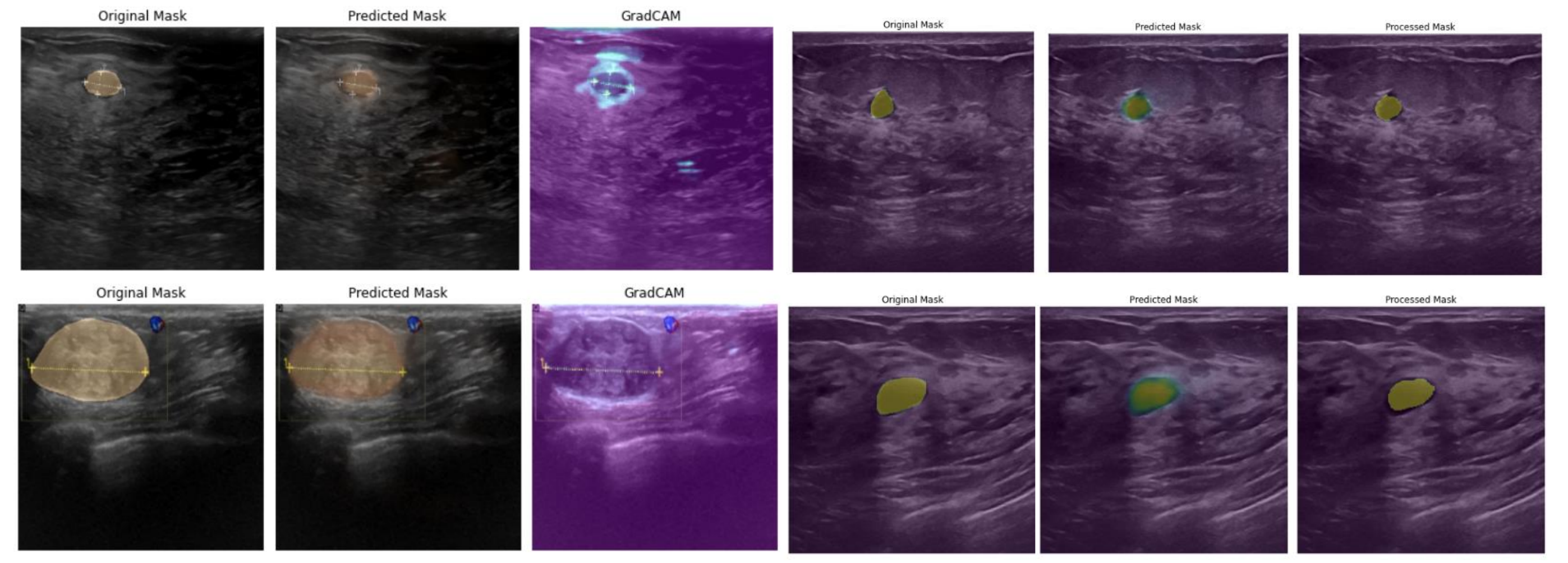}}
    \caption{Results of predicted, processed and GradCAM images}
    \label{fig:Aug}
\end{figure*}

\section{Experimental Results And Analysis}
This section deals with an experimental analysis and implementation of the proposed methodology. The implementation and the analysis of the proposed framework are further categorized with a top-down approach into the histogram equalization, U-Net-based segmentation and understanding of the XAI, etc. The implementation results and the analysis will be portrayed and explained simultaneously.

\subsection{Computational Specifications}

This subsection stipulates the hardware and software specifications required for the implementation. The entire implementation of the framework is performed in Windows. With the aid of other tools, such as  Kaggle API and CUDA, this OS facilitates establishing the neural network \cite{qi2022yolo5face}. The computational backend programs for all these tools are in the Python programming language. Moreover, these algorithms do not require a high-performance graphics processing unit. The specifications and minimum requirements for training and testing the framework are depicted in Table \ref{table1}.

\begin{table}[ht]
\caption{}
\label{table1}
\begin{center}
\scalebox{1}
{
\begin{tabular}{|c|c|}
\hline
\textbf{Specifications} & \textbf{System's Configuration}  \\ 
\hline
Operating system & Windows\\
\hline
CPU & Intel® i7 10th gen\\
\hline
RAM & 32 GB Usable\\
\hline
GPU & Nvidia Geforce 3070Ti\\
\hline
Frameworks & Tensorflow\\
\hline
\end{tabular}
}
\end{center}
\end{table}

\begin{figure}[ht]
    \centering
    \includegraphics[width=\linewidth]{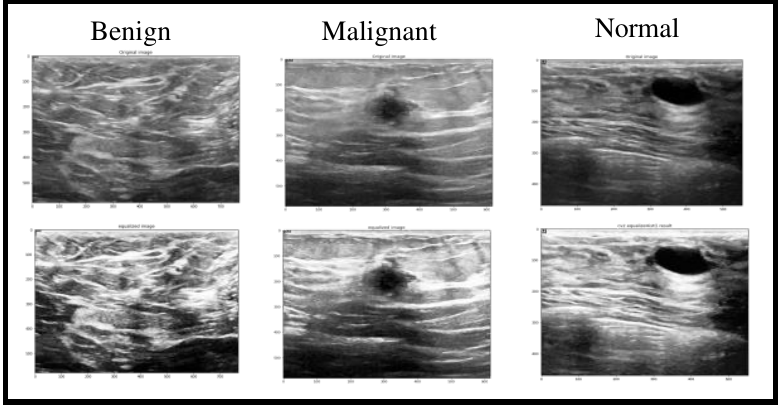}
    \caption{The Experimental Results of Histogram Equalization}
\end{figure}

\begin{figure*}
    \centering
    \centerline{\includegraphics[width=\linewidth]{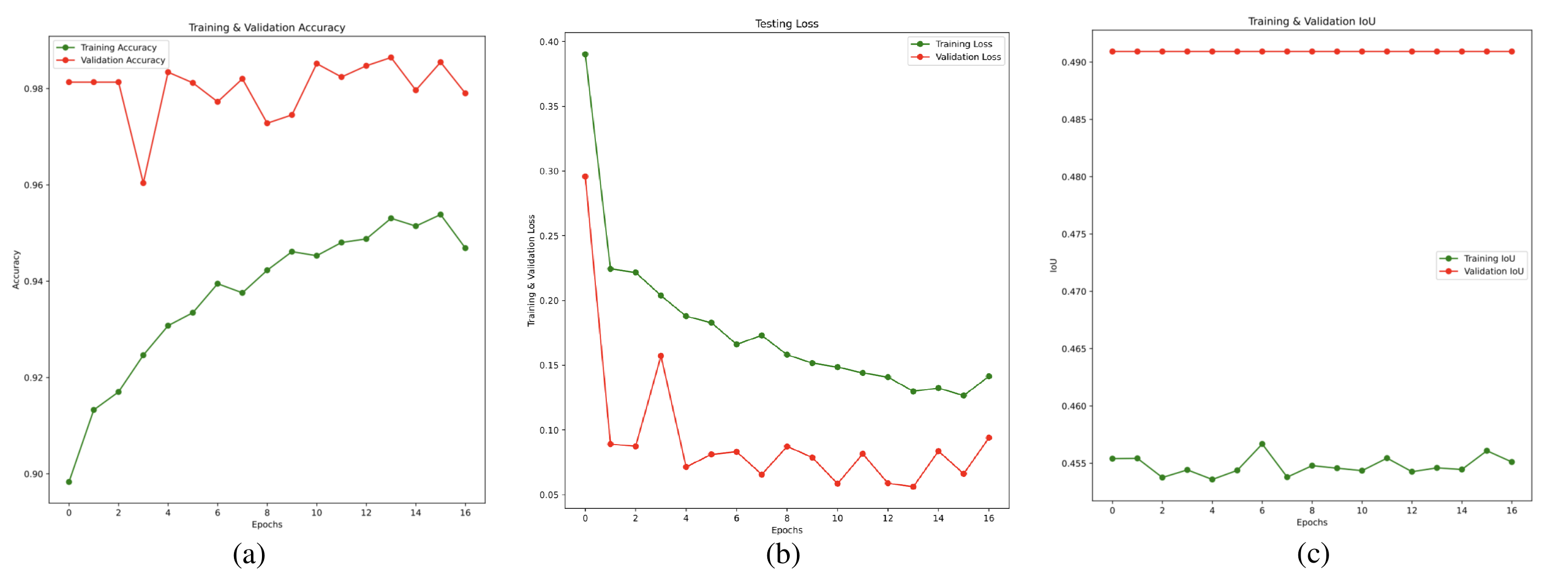}}
    \caption{U-net Training and Validation Learning Curves (a) Accuracy , (b) Loss \& (c) IoU}
        \label{fig:graphs}
\end{figure*}

\subsection{Evaluating trained Proposed architecture}
AttentionUNet/UNet is the best know model for Multi-Class Segmentation,  thats why we will be creating an Attention UNet Model. As the image Dataset is old, please do not use it for any new medical operations.

All the images are 500 X 500 pixels. RAM will not be enough so we will resize the Image to 256 X 256 pixels. In this experiment, the histogram equalisation is followed by training of the ViT model for breast picture segmentation. The U-Net architecuture will consist of an Encoder Block, Decoder Block and an Attention Gate. Our system is evaluated using the standard standards for segmentation model evaluation. Fig. depicts the accuracy and loss learning curves for model training and model validation. These learning plots show a well-fitted learning algorithm because both the validation and training curves maintain a stable point with little gap. Performance was increased by incorporating three tasks into the training of the effective UNet model \cite{9531646} at once: computing output, troubleshooting errors, and fine-tuning hyper-parameters. The maximum training and validation accuracies are 95\% and  99\%, respectively, after numerous iterations of tuning the hyper-parameters, which are accomplished with an idea. For Further analysis of the proposed architecture the appropriate experimental parameters have been taken into consideration, such as IoU, Loss, Accuracy, Precision, Recall, F1-Score, etc.

\begin{equation}
\text { loU score }(A, B)=\frac{\text { Area of Intersection }(A, B)}{\text { Area of Union }(A, B)}=\frac{A \cap B}{A \cup B}
\end{equation}

\subsubsection{Observations}
\begin{itemize}
    \item After 12 epochs model, the segmentation model results are outstanding.
    \item The model was quickly able to detect black round spots but failed when the shape was irregular (Not the case with the current model because it is trained with high Steps Per Epoch (SPE)).
    \item It also gets confused between the dark areas, which makes sense.
    \item Training in chunks of 20 Epochs; will deliver reasonable control over the model, and the model will also perform well.
    \item Surprisingly the results on Validation Data are way better than the Training Data on IoU. This may indicate that the model can perform way better than it can at the current point. The Loss is not Perfect. It increases in the last, but the model constructions look perfect.
\end{itemize}

%
\IEEEpeerreviewmaketitle

\section{Conclusion}
In this article, we investigated the use of microscopic hyperspectral imaging technology in breast tumour tissue microarrays and achieved breast tumour tissues' automated detection. We anticipate that it will help pathologists make a pathological diagnosis of breast cancer. The tissue microarray is convenient for creating a breast cancer data set in the experiment because of its modest size, considerable information richness, and one slice that contains the pathological condition of several patients. Additionally, the technology reduces the time required for sample preparation, data collecting, and picture processing, increasing the effectiveness of diagnostic procedures. The proposed framework combines the benefits of hyperspectral imaging technology and deep learning (feature learning and classification capabilities across multiple images) in terms of data processing. On a collection of photos that have been expertly labelled for a hyperspectral data set, conventional machine learning techniques and our approach are compared. The breast cancer nest tissue has been successfully captured using automatic feature extraction of U-Net, significantly enhancing classification accuracy. In order to explore more suitable methods for analysing microscopic hyperspectral pathological images, we are currently working on gathering more pathological samples and using the suggested method as a first step to quantitatively analyse the morphological characteristics of the breast cancer microenvironment.


\ifCLASSOPTIONcompsoc
  \section*{Acknowledgments}
\else
  \section*{Acknowledgment}
\fi

The International Institute of Information Technology, Naya Raipur, provided support and technical assistance for this study.



%
\bibliography{references}

\end{document}